\documentclass{article}

\usepackage{microtype}
\usepackage{graphicx}
\usepackage{subfigure}
\usepackage{booktabs} 

\usepackage{hyperref}

\usepackage[accepted]{icml2023}

\usepackage{amsmath}
\usepackage{amssymb}
\usepackage{mathtools}
\usepackage{amsthm}

\usepackage[capitalize,noabbrev]{cleveref}

\usepackage[utf8]{inputenc} 
\usepackage[T1]{fontenc}    
\usepackage{hyperref}       
\usepackage{url}            
\usepackage{booktabs}       
\usepackage{amsfonts}       
\usepackage{nicefrac}       
\usepackage{microtype}      
\usepackage{xcolor}         
\usepackage{amsmath,amssymb}
\usepackage{enumitem}
\usepackage[capitalize]{cleveref}
\usepackage{bbm}
\usepackage{wrapfig}
\usepackage[export]{adjustbox}
\usepackage{float}

\usepackage{tikz}
\usepackage{pgfplots}
\usetikzlibrary{fit}
\usetikzlibrary{arrows}
\usetikzlibrary{arrows.meta}
\usetikzlibrary{positioning}
\usetikzlibrary{matrix}
\usepackage{graphicx}
\usetikzlibrary{calc}
\usepackage{graphmod}

\input{m.macros}
\input{m.symbols}

\def\x{\vec{x}}

\def\setX{\set{X}}

\renewcommand\nn[1][n]{^{(#1)}}

\def\pthetap{P_{\theta_P}}
\def\pthetaq{Q_{\theta_Q}}

\newcommand\pthetazx[1][\x]{f_{\theta #1}}
\newcommand\pthetazw[1][\w]{f_{\theta #1}}
\newcommand\pthetax[1][\x]{F_{\theta #1}}
\newcommand\pthetaw[1][\w]{F_{\theta #1}}
\def\pxemp{P_{0x}}
\def\pwemp{P_{0w}}

\newcommand\XN[1][N]{\ensuremath{\mathbb{X}\nn[#1]}}
\newcommand\WN[1][N]{\ensuremath{\mathbb{W}\nn[#1]}}

\def\pthetaXNWN{\pr[\theta,\XN,\WN]}

\def\eff{\mathcal{F}}

\newcommand{\indep}{\perp \!\!\! \perp}

\usepackage{mbubble}
\mbubblestyle{ww}{font=\tiny,label=ww,color=teal}
\mbubblestyle{ms}{font=\tiny,label=ms,color=blue}
\theoremstyle{plain}

\theoremstyle{definition}

\theoremstyle{remark}

\icmltitlerunning{Latent Subgroup Shifts with High-dimensional Observations}

\begin{document}
\marginparsep=1mm

\twocolumn[
\icmltitle{Prediction under Latent Subgroup Shifts with High-Dimensional Observations}



\icmlsetsymbol{equal}{*}

\begin{icmlauthorlist}
\icmlauthor{William I. Walker}{gatsby}
\icmlauthor{Arthur Gretton}{gatsby}
\icmlauthor{Maneesh Sahani}{gatsby}
\end{icmlauthorlist}

\icmlaffiliation{gatsby}{Gatsby Computational Neuroscience Unit, UCL, London, UK}

\icmlcorrespondingauthor{William I. Walker}{william.walker.18@ucl.ac.uk}

\icmlkeywords{Machine Learning, ICML, Causal Learning, Generative Models}

\vskip 0.3in
]



\printAffiliationsAndNotice{}  

\begin{abstract}
We introduce a new approach to prediction in graphical models with latent-shift adaptation, i.e., where source and target environments differ in the distribution of an unobserved confounding latent variable.
Previous work has shown that as long as "concept" and  "proxy" variables with appropriate dependence are observed in the source environment, the latent-associated distributional changes can be identified, and target predictions adapted accurately.  
However, practical estimation methods do not scale well when the observations are complex and high-dimensional, even if the confounding latent is categorical.
Here we build upon a recently proposed probabilistic unsupervised learning framework, the recognition-parametrised model (RPM), to recover low-dimensional, discrete latents from image observations.
Applied to the problem of latent shifts, our novel form of RPM identifies causal latent structure in the source environment, and adapts properly to predict in the target.
We demonstrate results in settings where predictor and proxy are high-dimensional images, a context to which previous methods fail to scale.
\end{abstract}

\section{Introduction}

In real world prediction problems, challenges often arise owing to shifts between training and test distributions.
%
%
%
For example, suppose we want to predict language exam performance ($Y$) given past written essays ($X$) from the same students. 
A naive model trained on students from one school may fail to accurately predict the performance of students from another where the mix of socio-economic backgrounds differs, as these contextual factors can affect not only the marginal distributions of prior and future scores, but also the relationship between them.
%
Attempts to mitigate the impact of such distribution changes on performance of tasks such as classification depend on assumptions about which distributions have changed and which probabilities are conserved. 
In covariate shift \cite{SHIMODAIRA2000227}, the distribution of the observed covariate $X$ changes but the conditional probability of $Y|X$ remains the same. 
In the label shift setting \cite{gart}, the distribution of the predicted variable $Y$ shifts but the conditional dependence of $X|Y$ is preserved. 
%
However, in our school example, both the distributions on $X$ and $Y$ as well as their conditionals may shift, and these assumptions are not general enough.

Latent shift assumes a change in the distribution of an unobserved confounding latent subgroup variable $U$ that influences both $X$ and $Y|X$, while the conditional dependence of $X$, $Y$ and any other observed variables on $U$ are preserved.
\citet{adapting2023} introduce a graph with proxy and concept variables observable in the source environment, which allows the identification of the distribution of $U$ and its confounding effects.  They show that this model can be used to adapt predictions in the target environment to shifts in the distribution of $U$.

However, the learning algorithm they propose has difficulty scaling to settings where the observed covariates are high-dimensional and complex, such as images or documents.
This is a setting in which the Recognition-Parametrised Model (RPM) \cite{walker2023} is helpful.  
The RPM makes it possible to learn a tractable underlying latent graphical structure from high dimensional observations, without needing an explicit generative model.
Here we propose a novel adaptation of the RPM in which factors of the graph involving high dimensional observations are parametrised in recognition form, while keeping generative factors for categorical variables including the predicted variable $Y$.
We refer to this model as a "partial" RPM.  
We show that it can be used to identify the confounding latent within a source distribution $P$ with complex observed variables, and to adapt for latent subgroup shift in a target distribution $Q$, outperforming previous models as the dimensionality of the observation increases.
We demonstrate an application in which the observations are images from the CIFAR-10 dataset, and quantify predictive performance after shift for different observed sample sizes in source and target.

\section{Previous Work}

Many recent studies have sought to learn models that adapt to shifts in distribution from source $P$ to target $Q$, with different assumptions on which conditional probabilities are preserved.
%
Covariate shift \cite{SHIMODAIRA2000227,covariate2,covariate3,covariate4,covariate5,covariate6,covariate7,covariate8} considers the case in which $P(X) \neq Q(X)$ but the conditional probability $P(Y|X)=Q(Y|X)$ stays the same.
This is usually solved by reweighting the classifier loss by $Q(X)/P(X)$ in an attempt to make the source data look like it was drawn from the target data.
Similarly, label shift \cite{label1,label2,label3,label4,label5,label6,label7,label8,label9,label10,label11,label12,label13} considers the case where $P(Y) \neq Q(Y)$ but the conditional probabilities $P(X|Y)=Q(X|Y)$ remain the same.
Again, this shift is usually mitigated by reweighting the classifier loss, this time with $Q(Y)/P(Y)$.
Both the problem settings above have been framed as assumptions on the causal structure of the generation of the data \cite{causalLink}.

Latent shift was framed in the causal view by \citet{yue}, where a shift in distribution of an unobserved latent confounder changes the distribution on both $X$ and $Y$.
This work finds a proxy for $X$ and one for $Y$, both caused by $U$, and finds an invariant bridge function to remove the effect of the latent shift.
However, as pointed out by \citet{adapting2023}, this approach does not guarantee that the mapping from $U$ to either proxy is identifiable, a condition which is necessary to correctly adapt to the shift in distribution of $U$.
\citet{adapting2023} solve the identifiability problem of $U$ by introducing proxy variable $W$ and concept variable $C$ as in the graph of \cref{fig:graph}.
Both $C$ and $W$ are observed in source $P$ but not in target $Q$.
They then introduce an algorithm to adapt to shifts in the latent distribution under the assumption that all probabilities conditioned on $U$ are preserved between source and target.
However, as we demonstrate here, this algorithm does not scale well as the dimensionality of $X$ and $W$ grow, although such high-dimensional covariates may be encountered in realistic settings.  
Our goal in this study is to develop a new approach to estimation and prediction in the latent-shift setting that scales to real-world dimensionalities

\section{Latent Shift Adaptation}

\begin{figure}
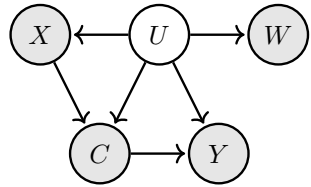

\begin{center}
\graphmod{graphical_models/graphical_model}[genconf]
\end{center}
\caption{Graphical model of observations and confounding latent $U$. $X$ is the observed variable, $C$ is the concept variable, $W$ is the proxy variable, and $Y$ is the predicted variable. All of $\{X,C,W,Y\}$ are observed in the source, but only $X$ is available in the target.}
\label{fig:graph} 
\end{figure}

We follow \citet{adapting2023} by assuming the graphical model of \cref{fig:graph}, where $U$ is the unobserved latent, $X$ is the observed covariate, $Y$ the variable to be predicted, $W$ is the proxy variable, and $C$ is a concept variable.
The concept $C$ ensures that $Y \indep X |C,U$, while the proxy $W$ provides sufficient additional information about $U$ to make it identifiable.
%
Further discussion about the inclusion of $C$ and $W$ can be found in  \cite{adapting2023}.
We take $C$, $Y$, and $U$ to be discrete variables whereas $X$ and $W$ may be discrete or continuous, possibly vector-valued and high-dimensional.
Our goal is to predict $Y$ given observations $X$ from the target distribution $Q$, having trained on observations $\{X,C,W,Y\}$ from the source distribution $P$.
We assume the distribution of the unobserved latent $U$ can shift between $P$ and $Q$, but that all probabilities conditioned on $U$ are preserved across the source and target.

\subsection{The Partial Recognition-Parametrised Model}

Let $X$ and $W$ be (discrete or continuous) vector-valued, observed random variables and $C$ and $Y$ be discrete observed random variables.
We want to identify the underlying discrete latent variables $U$ giving rise to the conditional independencies shown in the graph of \cref{fig:graph}. 
The conditional dependence structure implies the factorisation
\begin{align*}
    p(U,\!C,\!Y,\!X,\!W) &= p(X|U)p(W|U)p(C|X,\!U)p(Y|C,\!U)p(U)
\end{align*}
The Recognition-Parametrised Model \citep[RPM;][]{walker2023} makes it possible to learn a flexible recognition model (e.g.\ a neural network) without an explicit generative model for the observations.
This is appropriate to the current setting, where there is no need to generate either $X$ or $W$ in either source or target setting. 
Therefore, we introduce RPM-like terms in place of the conditional dependencies $p(X|U)$ and $p(W|U)$
\begin{eqnarray}
    p(X|U) &\rightarrow& \frac{\pxemp(X)\pthetazx(U| X)}{\pthetax(U)} \nonumber \\
    p(W|U) &\rightarrow& \frac{\pwemp(W)\pthetazw(U| W)}{\pthetaw(U)}
\end{eqnarray}
$\pxemp(X) = \frac 1N\sum_n \delta(X - X\nn)$ is the empirical measure with atoms at the $N$ data points $X\nn$. Similarly, $\pwemp(W) = \frac 1N\sum_n \delta(W - W\nn)$.
The factor $\pthetazx(U| X)$ is a parametrised recognition distribution and $\pthetax(U)=\intdx[dX] \pxemp(X)\pthetazx(U | X)$ is the mixture with respect to $\pxemp$.  The factors $\pthetazw$ and $\pthetaw$ are defined analogously.
Together, these terms define a normalised joint model 
\begin{multline}\label{eq:joint}
   \pthetaXNWN{U,C,Y,X, W}=P_{\phi}(C|X,U)P_{\psi}(Y|C,U)\\ \pthetap(U)
   \frac{\pxemp(X)\pthetazx(U| X)}{\pthetax(U)} \frac{\pwemp(W)\pthetazw(U| W)}{\pthetaw(U)} 
\end{multline}
where $\theta_x$, $\theta_w$, $\theta_p$, $\phi$, and $\psi$ are learned parameters.
$\pthetap(U)$ is a normalised distribution on the latent.
The observed datasets $\XN = \{\setX\nn[1] \dots \setX\nn[N]\}$ and $\WN$ appear in the subscript to highlight that the model parametrisation depends on these data through $\pxemp$ and $\pwemp$ and the mixtures.

\subsection{RPM Latent Adaptation}\label{sec:algorithm}
We observe data $\{X,Y,C,W\}$ from source $P$, and $\{X\}$ from target $Q$ and seek to estimate $Q(Y|X)$.
The steps for learning and adaptation in our model are detailed below.

\paragraph{Step 1: Learn RPM for source P.}

The parameters $\{\theta_x, \theta_w, \theta_u, \phi, \psi\}$ of the source model of \cref{eq:joint} are learnt by maximizing the free energy using Expectation-Maximization (EM).
\begin{align}\label{eqn:P_free_energy}
    \eff &\underset{+C}{=} \langle \log \pthetazx(U| X) \rangle_{\eta(U)} - \langle \log \pthetax(U) \rangle_{\eta(U)} \nonumber \\
    & + \langle \log \pthetazw(U| W) \rangle_{\eta(U)} - \langle \log \pthetaw(U) \rangle_{\eta(U)} \nonumber \\
    & + \langle \log P_{\phi}(C|X,U)\rangle_{\eta(U)} + \langle \log P_{\psi}(Y|C,U)\rangle_{\eta(U)} \nonumber \\
    & + \langle \log \pthetap(U) \rangle_{\eta(U)} + H[\eta(U)]
\end{align}
where $\eta(U)$ is the variational distribution and $H$ is the entropy function.
The expectation step to update $\eta(U)$ in EM is exact for discrete  $U$  \cite{walker2023} and so EM will converege to a (local) maximum of the likelihood.

\paragraph{Step 2: Learn the shifted prior $\pthetaq(U)$ on target.}

Now only using $\{X\}$ from target $Q$, we learn the joint $Q(U,X)$ by maximizing the free energy. 
Note that any conditional probability dependent on $U$ remains the same as in $P$. Only the prior changes: $P(U) \rightarrow Q(U)$.
Thus, we can reuse parameters learned from the source distribution. We then have: 
\begin{align}
Q(U|X) &= \frac{Q(X|U)\pthetaq(U)}{Q(X)} \nonumber \\
&= \frac{P(X|U)\pthetaq(U)}{Q(X)} \nonumber \\
&\propto \frac{\pthetazx(U| X)}{\pthetax(U)} \pthetaq(U)
\end{align}
Now we maximize the free energy to learn the parameter $\theta_Q$ while keeping all other parameters fixed
\begin{align}
    \eff &\underset{+C}{=} \langle \log \pthetazx(U| X) \rangle_{\nu(U)} - \langle \log \pthetax(U) \rangle_{\nu(U)} \nonumber \\
    & + \langle \log \pthetaq(U) \rangle_{\nu(U)} + H[\nu(U)]
\end{align}
where $\nu(U)$ is the variational distribution which can be computed analytically in the expectation step of EM.

\paragraph{Step 3: Make predictions $Q(Y|X)$.}
Using all the factors learned in the previous steps, make predictions
\begin{align}
    Q(Y|X) 
  &= \sum_{U,C} Q(Y|C,U) Q(C|U,X) Q(U|X) \nonumber \\
  &= \sum_{U,C} P_{\psi}(Y|C,U) P_{\phi}(C|U,X) Q(U|X)
\end{align}
where again we use the fact that by conditioning on $U$, we can use the probabilities learned on the source P.

\section{Experiments}

\subsection{Scaling with observation dimensionality}

\begin{figure}
\begin{center}
\includegraphics[width= 1 \linewidth, trim= 0cm 0cm 0cm 2cm, clip]{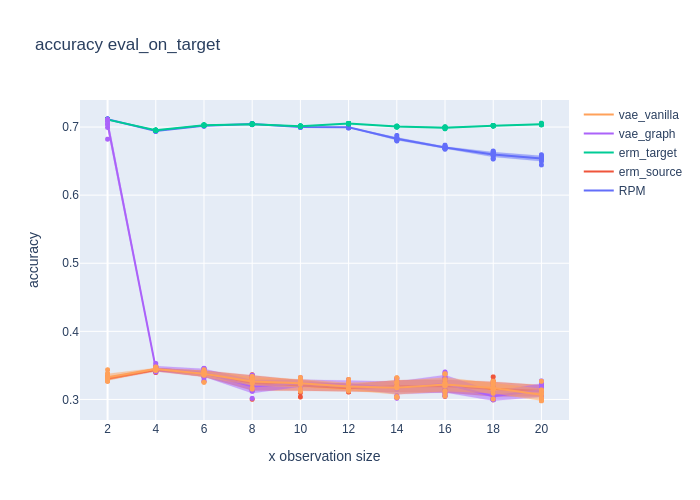}
\vspace{-1.0cm} 
\end{center}
\caption{Accuracy in predicting $Y$ given $X$ from the target distribution, as a function of the dimensionality of $X$. erm\_source predicts using a model trained on source distribution $X$ and $Y$, giving a lower bound on prediction performance. erm\_target predicts using an oracle model trained on target distribution $X$ and $Y$, giving an upper bound on prediction performance.}
\label{fig:obs_dim}
\end{figure}

To show how the partial RPM is able to handle complex observations, we used a simulated numerical setting with varying dimensionalities of $X$. 

We constructed data with binary $U$, $Y$, $W$, and $C$ but continuous $X$. 
In the source distribution $P$, $U$ was drawn with $P(U=1)=0.9$ and in the target $Q$, $Q(U=1)=0.1$. 
The first two dimensions of $X$ were drawn from a Gaussian with mean dependent on $U$. 
All the other dimensions of $X$ were simply $\{10,-10\}$ with probability 0.5 and had no relevance to the prediction of $Y$. 
This structure was chosen based on the hypothesis that previous methods using the structured Wassertein Autoencoder (WAE) or Variational Autoencoder (VAE) \cite{VAE}, which must learn to reconstruct the observation $X$, would be challenged as  irrelevant dimensions of $X$ were added and so learn a worse representation of $U$.
Further details of the generative model are given in Appendix \ref{apdx:x_dim}.

We compared our approach to several baselines.
Empirical risk minimization is a neural network (NN) trained on source data (erm\_source) to predict $Y$ from $X$. 
It is then simply applied to target data $X$ and serves as a lower bound to prediction.
Conversely, erm\_target is an oracle network trained on \emph{target} data $X$ and $Y$ and is an upper bound to prediction performance.
%
%
Latent shift adaptation using the structured WAE (vae\_graph) \cite{adapting2023} is trained with a recognition model $\hat{p}_e(\Tilde{U}|X,C,Y,W)$ and a decoder that respects the conditional independence structure of the model and reconstructs all observations by $\hat{Y}=f_y(C,U)$, $\hat{C}=f_c(X,U)$, $\hat{X}=f_x(U)$, and $\hat{W}=f_w(U)$.
Details of its training can be found in the Appendix \ref{apdx:x_dim}.
Latent shift adaptation using the vanilla VAE (vae\_vanilla) \cite{adapting2023} is trained with the same recognition model to $U$ $\hat{p}_e(\Tilde{U}|X,C,Y,W)$ but the decoder has no structure $\hat{p}_d(X,C,Y,W|\Tilde{U})$.
The partial RPM recognition model from $X$ to $U$ is a multilayer perceptron (MLP) with a single hidden layer of 100 units and ReLU activations, the same recognition model used in vae\_graph and vae\_vanilla, and the same as the NN in erm\_source and erm\_target. 
The partial RPM recognition model has binary $U$ as in the true generative model, unlike vae\_graph and vae\_vanilla where $U$ is allowed 5 categories. 
This is needed in the VAE models to capture more structure in $X$, and using binary $U$ in those cases results in worse performance.

As seen in \cref{fig:obs_dim}, partial RPM accuracy remains high as more dimensions are incorporated into $X$, while the vae\_graph performance falls to the erm\_source baseline.
This is consistent with our hypothesis that VAE models, unlike the RPM, require the latent $U$ to capture enough information to reconstruct observation $X$ and are thus affected by the noisy dimensions.
The RPM avoids this by dispensing with generation, and so focusing only on the information in $X$ that is shared with the other observed variables through $U$.
In the next section, we consider the case where $W$ is also continuous and high-dimensional.

\subsection{Latent shift with CIFAR-10 image observations}

\begin{figure}
\begin{center}
\includegraphics[width= 1 \linewidth, trim= 0cm 0cm 0cm 2cm, clip]{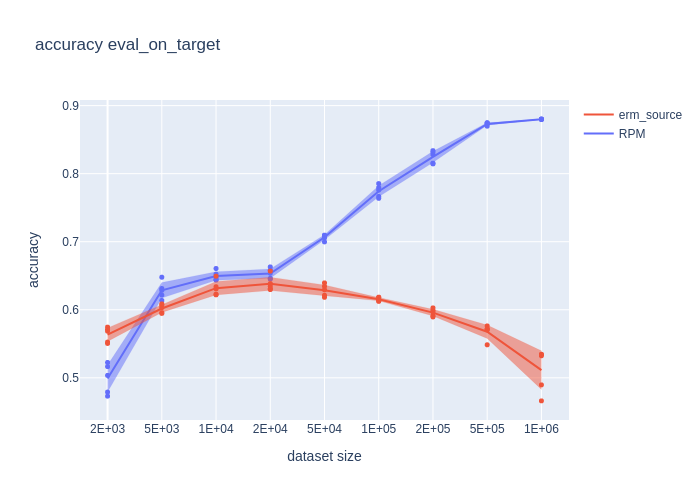}
\vspace{-1.0cm} 
\end{center}
\caption{Accuracy in predicting $Y$ given $X$ from the target distribution. erm\_source predicts using a model trained on source distribution $X$ images and $Y$ and does not adapt to latent shift.}
\label{fig:cifar}
\end{figure}

To demonstrate close to real-world scaling, we allow $X$ and $W$ to be images from different sets of classes in the CIFAR-10 dataset.
The data are generated with $U$, $Y$, $C$, $\Tilde{X}$, and $\Tilde{W}$ discrete assuming sizes of $k_u=3$, $k_y=2$, $k_c=3$, $k_{\Tilde{X}}=2$, and $k_{\Tilde{W}}=3$ respectively. 
$U$ is drawn from the prior according to $P(U)$ for the source and $Q(U)$ for the target.
$P(Y|C,U)$, $P(C|\Tilde{X},U)$, $P(\Tilde{X}|U)$, and $P(\Tilde{W}|U)$ are all drawn according to probability matrices given in Appendix \ref{apdx:cifar}.
The variables $\Tilde{X}$ and $\Tilde{W}$ give the class identity of the CIFAR-10 image for $X$ and $W$, which are then randomly drawn from the corresponding class.
We chose $X$ and $W$ from a set of classes that did not overlap.
Although $X$ and $W$ are similar here, this is not essential to the RPM model.
As $\pthetazx(U| X)$ and $\pthetazw(U| W)$ are parameterized by two separate neural networks (NN), the observations can be completely different in dimensionality or statistical structure \cite{walker2023}.

In the partial RPM, $\pthetazx(U| X)$ and $\pthetazw(U| W)$ are convolutional NNs that output discrete probabilities on $U$ and $P_{\phi}(C|X,U)$ is a convolutional NN to discrete probabilities on $C$. 
As all terms are discrete in $P_{\psi}(Y|C,U)$, its parameters are simply the corresponding probability matrices.
Learning is achieved by maximizing the free energy in \cref{eqn:P_free_energy},  with adaptation following by the steps elaborated in \cref{sec:algorithm}.
In \cref{fig:cifar}, the target adaptation performance is shown as a function of source and target dataset size.
The partial RPM (blue) is compared to erm\_source (red) with a NN that is the same size as the RPM recognition network for $\pthetazx(U| X)$ but with output size $k_y$.
erm\_source demonstrates that no matter the size of the dataset, simple prediction trained on source will not account for the shift and perform poorly on target.
We see here that the partial RPM can adapt to the latent shift even for complex inputs such as CIFAR-10 images with sufficient training data.

\section{Conclusion}

We have introduced a new method for latent shift adaptation using the RPM.
Our method scales to high-dimensional, structured data by avoiding the need to reconstruct that data.
We have shown that previous models do not perform well as the dimensionality of the observed covariate is increased, even when the added dimensions only contain noise, while the RPM approach is still able to predict with these observations.
Furthermore, we demonstrated that the observations can be highly structured as in CIFAR-10 dateset and the RPM still adapts to latent shifts with enough training data.
This makes the RPM a strong model for real world applications.

\bibliography{icml_workshop}
\bibliographystyle{icml2023}

\newpage
\appendix
\onecolumn
\section{Numerical Simulation Experiment} \label{apdx:x_dim}

We describe the numerical experiments in which we change the dimensionality of the observation $X$. $U$, $Y$, $C$, and $W$ are discrete while $X$ is continuous.  We let the dimensions of each of the variables to be $k_U=2$, $k_C=2$, $k_Y=2$, and $k_W=2$. The dimensionality of $X$ is varied from 2 to 20 in steps of 2. Below is how each variable is generated, closely following \cite{adapting2023}. $\Vec{o}(U)$ is the one-hot representation of the variable $U$. $I_{k_X}$ is the identity matrix of size $k_X \times k_X$. $\Theta$ is the Heaviside function, $\Theta(z)=1$ if $z > 0$ and 0 otherwise.

\begin{eqnarray}
    U &\sim& \textrm{Categorical}(\boldsymbol{\pi}) \nonumber \\
    W|U=u &\sim& \Theta(\mathcal{N}(\Vec{o}(U)\boldsymbol{M}_{W|U}, 1)) \nonumber \\
    X[0:2]|U=u &\sim& \mathcal{N}(\Vec{o}(U)\boldsymbol{M}_{X|U}, \boldsymbol{I}_{k_X}) \nonumber \\
    X[2:] &\sim& 10 \times(2\times\textrm{Bernoulli}(0.5) -1)\nonumber \\
    C|X=x,U=u &\sim& \textrm{Bernoulli}(\textrm{logit}^{-1}(X\boldsymbol{M}_{C|X,U=u} + \Vec{o}(U)\boldsymbol{M}_{C|U})) \nonumber \\
    Y|C=c,U=u &\sim& \textrm{Bernoulli}(\textrm{logit}^{-1}(C\boldsymbol{M}_{Y|C,U=u} + \Vec{o}(U)\boldsymbol{M}_{Y|U})) \nonumber
\end{eqnarray}

Here are each of the generative parameters:
\begin{eqnarray}
\boldsymbol{M}_{W|U} =
\begin{bmatrix}
-3 & 3
\end{bmatrix}^T, 
\boldsymbol{M}_{X|U} = 
\begin{bmatrix}
-0.5 & 0.5\\
0.5 & -0.5
\end{bmatrix}, 
\boldsymbol{M}_{C|U} = 
\begin{bmatrix}
-1 & 1
\end{bmatrix}^T
\nonumber \\
\boldsymbol{M}_{C|X,U=u_0} = 
\begin{bmatrix}
-1 & 1
\end{bmatrix}, 
\boldsymbol{M}_{C|X,U=u_1} = 
\begin{bmatrix}
1 & -1
\end{bmatrix}
\nonumber \\
\boldsymbol{M}_{Y|U} = 
\begin{bmatrix}
-2 & 2
\end{bmatrix}^T, 
\boldsymbol{M}_{Y|C,U=u_0} = 
\begin{bmatrix}
-1 & 1
\end{bmatrix}^T, 
\boldsymbol{M}_{Y|C,U=u_1} = 
\begin{bmatrix}
1 & -1
\end{bmatrix}^T
\end{eqnarray}
For the source distribution, $\boldsymbol{\pi} = [0.1, 0.9]$ while in the target distribution $\boldsymbol{\pi} = [0.9, 0.1]$. We used $7\times 10^4$ samples of both source and target distributions.

We chose these parameters such that any VAE model would struggle to reconstruct the data. To see this, notice that the latent dimensionality $k_U$ is fixed at 2. The first two dimensions of $X$ contain all the relevant information about $U$ and all other dimensions are noise. However, the noisy dimensions have a large scale. Therefore, when the WAE reconstructs the data X, error in the noisy dimensions contribute significantly to the loss. Therefore, the WAE may learn to represent the noisy dimensions in the latent $U$ and ignoring the parts of $X$ that pertain to the latent shift. As the dimensionality of $X$ grows, this problem only worsens. In the RPM, however, there is no reconstruction in the loss term and therefore no need to represent the noisy dimensions. The $U$ that maximizes the loss is only that which makes $X$ conditionally independent of the rest of the graph.

The ERM source and target models both were MLPs with one hidden layer of 100 units with ReLU activations. They were trained for 500 epochs with batch size of 128 and learning rate of 0.01 in stochastic gradient descent (SGD). This learning rate was rediced by a factor of 10 if the training loss did not decrease by 0.01 in the last 20 epochs. The minimum learning rate was set at $10^{-7}$. Weight decay of $10^{-6}$ was used.

The WAE and VAE approaches used a latent of dimensionality 5. The encoder models both were MLPs with one hidden layer of 100 units with ReLU activations. The decoder for WAE used the factored form of the joint distribution implied by the graphical model of \cref{fig:graph}. The separate decoder networks $\{f_Y,f_C,f_X,f_W\}$ all were MLPs with one hidden layer of 100 units with ReLU activations. Details on the loss function are found in \cite{adapting2023}. The WAE was fit with RMSprop for 500 epochs using $10^-4$ learning rate. The annealing strategy was the same as the ERM models.

The partial RPM used a latent of dimensionality 2. The recognition model from $X$ to $U$ ($\pthetazx(U| X)$) and the generative model from $X$ and $U$ to $C$ ($P_{\phi}(C|X,U)$) were MLPs with one hidden layer of 100 units with ReLU activations. $\pthetazw(U| W)$ and $P_{\psi}(Y|C,U)$ were simply  probability matrices. All parameters were learned using Adam for 2000 epochs with learning rate of $10^{-3}$ and batch size of 1000. We also multiply the entropy in the loss by a factor of 5.0 which linearly reduces to 1.0 in 400 epochs.

We ran 10 random initializations of each model on the same generated data for every size of observation $X$.

\section{CIFAR-10 Observation Experiment} \label{apdx:cifar}

In the CIFAR-10 experiments, we make observed $X$ and $W$ both CIFAR-10 images. The generative process for these observations is described here.

Here, $U$, $Y$, and $C$ are discrete. Let $\Tilde{X}$ and $\Tilde{W}$ be discrete variables that determine the class from which CIFAR-10 images for $X$ and $W$ are randomly drawn. We let the dimensionalities of each of the variables be $k_U=3$, $k_C=3$, $k_Y=2$, $k_{\Tilde{X}}=2$, and $k_{\Tilde{W}}=3$. $\Vec{o}(v)$ is the one-hot representation of some variable $v\in V$. 

\begin{eqnarray}
    U &\sim& \textrm{Categorical}(\boldsymbol{\pi}) \nonumber \\
    \Tilde{W}|U=u &\sim& \textrm{softmax}(\boldsymbol{M}_{\Tilde{W}|U}\Vec{o}(U)) \nonumber \\
    \Tilde{X}|U=u &\sim& \textrm{softmax}(\boldsymbol{M}_{\Tilde{X}|U}\Vec{o}(U)) \nonumber \\
    C|\Tilde{X}=\Tilde{X},U=u &\sim& \textrm{softmax}(\boldsymbol{M}_{C|\Tilde{X}}\Vec{o}(\Tilde{X}) + \boldsymbol{M}_{C|U}\Vec{o}(U)) \nonumber \\ 
    Y|C=c,U=u &\sim& \textrm{sigmoid}(\boldsymbol{M}_{Y|C}\Vec{o}(C) + \boldsymbol{M}_{Y|U}\Vec{o}(U)) \nonumber 
\end{eqnarray}

Here are each of the generative parameters:
\begin{eqnarray}
\boldsymbol{M}_{\Tilde{W}|U} =
\begin{bmatrix}
1e2 & -1e20 & -1e20 \\
-1e20 & 1e2 & -1e20 \\
-1e20 & -1e20 & 1e2 
\end{bmatrix}, 
\boldsymbol{M}_{\Tilde{X}|U} = 
\begin{bmatrix}
1e2 & 1e2 & -1e20 \\
1e2 & -1e20 & 1e2
\end{bmatrix}, 
\boldsymbol{M}_{C|U} = 
\begin{bmatrix}
5.0 & 5.0 & 0.5 \\
5.0 & 0.5 & 5.0 \\
0.5 & 5.0 & 5.0
\end{bmatrix}
\nonumber \\
\boldsymbol{M}_{C|\Tilde{X}} = 
\begin{bmatrix}
5.0 & 0.5 \\ 
5.0 & 5.0 \\
0.5 & 5.0
\end{bmatrix}, 
\boldsymbol{M}_{Y|U} = 
\begin{bmatrix}
5.0 & 5.0 & 0.5 \\
0.5 & 5.0 & 5.0
\end{bmatrix}, 
\boldsymbol{M}_{Y|C} = 
\begin{bmatrix}
5.0 & 5.0 & 0.5 \\
0.5 & 5.0 & 5.0
\end{bmatrix}
\end{eqnarray}
For the source distribution, $\boldsymbol{\pi} = \textrm{softmax}([1.0,0.1,0.1])$ while in the target distribution $\boldsymbol{\pi} = \textrm{softmax}([0.1,0.1,1.0])$. The prediction accuracy of $Y$ after latent shift was tested for 10 random initializations for each dataset, the size of which was chosen from $[2\times10^3, 5\times10^3, 10^4, 2\times10^4, 5\times10^4, 10^5, 2\times10^5, 5\times10^5, 10^6]$. The dataset size for both the source P and the target Q was the same.

The ERM source model was a convolutional NN. The first 2 layers were convolutional layers, with 3 input to 10 output channels and 10 input to 20 output channels. Each had kernel size 5 and the output was put through a 2D max pooling then ReLU function before the next layer. Then one linear layer of size 500 to 50 with ReLU activation and then one more linear layer to output. The output were logits on the discrete Y categories. It was trained for 5000 epochs with batch size of 2000 and learning rate of 0.001 using Adam.

The partial RPM used a recognition model for both $X$ ($\pthetazx(U| X)$) and $W$ ($\pthetazw(U| W)$) which was the same convolutional NN described for the ERM source model. The recognition model the generative model from $X$ and $U$ to $C$ ($P_{\phi}(C|X,U)$) was also the same convolutional NN. $P_{\psi}(Y|C,U)$ was a probability matrix. All parameters were learned using Adam for 20000 epochs with learning rate of $10^{-3}$ for the recognition parameters, $10^{-4}$ for the generative parameters, and $10^{-3}$ for the prior parameters. The batch size was 2000. We also multiply the entropy in the loss by a factor of 5.0 which linearly reduces to 1.0 in 3000 epochs.


\end{document}